\pdfoutput=1

\documentclass[11pt]{article}

\usepackage[usenames,dvipsnames,table]{xcolor}
\usepackage{emnlp2021}
\usepackage{times} 
\usepackage{latexsym}

\usepackage[T1]{fontenc}

\usepackage[utf8]{inputenc}

\usepackage{microtype}

\usepackage{amsfonts}
\usepackage{amsmath}
\usepackage{amssymb}
\usepackage{amsthm}
\usepackage{booktabs}
\usepackage{dsfont}
\usepackage{paralist}
\usepackage{times}
\usepackage{xspace}

\usepackage{multirow}
\usepackage{graphicx}
\usepackage{float} 

\usepackage{lipsum}
\usepackage{appendix}
\usepackage{enumitem}
\usepackage{pifont}
\usepackage{makecell}
\usepackage{caption}
\usepackage{latexsym}

\usepackage[usenames,dvipsnames,table]{xcolor}
\usepackage{soul}

\newcommand{\matr}[1]{\mathbf{#1}} 
\DeclareMathOperator*{\argmax}{arg\,max}

\newcommand{\ra}[1]{\renewcommand{\arraystretch}{#1}} 
\newcommand{\metric}{$\delta_{p,q}^\text{contr}$\xspace}

\newcommand{\draftonly}[1]{#1}
\renewcommand{\draftonly}[1]{}    
\newcommand{\draftcomment}[1]{\draftonly{#1}}

\newcommand{\alon}[1]{\draftcomment{{\color{blue}[#1]$_{aj}$}}}

\newcommand{\swabha}[1]{\draftcomment{{\color{cyan}[#1]$_{ss}$}}}
\newcommand{\ye}[1]{\draftcomment{{\color{red}[#1]$_{ye}$}}}

\newcommand{\cmark}{\ding{51}}%

\newcommand{\seq}[1]{\mathbf{#1}}

\DeclareRobustCommand{\hlmost}{{}}
\DeclareRobustCommand{\hlneg}{{\cellcolor{Lavender}}}
\DeclareRobustCommand{\hlleast}{{}}

\newcommand{\RNum}[1]{\uppercase\expandafter{\romannumeral #1\relax}}

\title{Contrastive Explanations for Model Interpretability}

\author{Alon Jacovi$^\heartsuit$\thanks{ \ \ Work done during an internship at the Allen Institute for Artificial Intelligence.} \;\;\;\; Swabha Swayamdipta$^\clubsuit$ \;\;\;\; Shauli Ravfogel$^\heartsuit$$^\clubsuit$ \\ \textbf{Yanai Elazar$^\heartsuit$$^\clubsuit$ \;\;\;\; Yejin Choi$^\diamondsuit$$^\clubsuit$ \;\;\;\; Yoav Goldberg$^\heartsuit$$^\clubsuit$}\\
$^\heartsuit$Bar Ilan University  \\
$^\clubsuit$Allen Institute for Artificial Intelligence \\
$^\diamondsuit$Paul G. Allen School of Computer Science and Engineering, University of Washington \\
  {\normalsize\tt  alonjacovi@gmail.com} \\
  {\normalsize\tt  \{swabhas,shaulir,yanaie,yoavg,yejinc\}@allenai.org}
  }

\begin{document}

\maketitle
\begin{abstract}
Contrastive explanations clarify why an event occurred \textit{in contrast to} another.
They are inherently intuitive to humans to both produce and comprehend. 
We propose a method to produce contrastive explanations in the latent space, via a projection of the input representation, such that only the features that differentiate two potential decisions are captured.
Our modification allows model behavior to consider only contrastive reasoning, and uncover which aspects of the input are useful for and against particular decisions.
Additionally, for a given input feature, our contrastive explanations can answer for which label, \textit{and against which alternative label}, is the feature useful.
We produce contrastive explanations via both high-level abstract concept attribution and low-level input token/span attribution for two NLP classification benchmarks.
Our findings demonstrate the ability of label-contrastive explanations to provide fine-grained interpretability of model decisions.\footnote{Code and data are available at \url{https://github.com/allenai/contrastive-explanations}.} 
\end{abstract}

\begin{figure}[t!]
\centering
\includegraphics[width=0.99\linewidth]{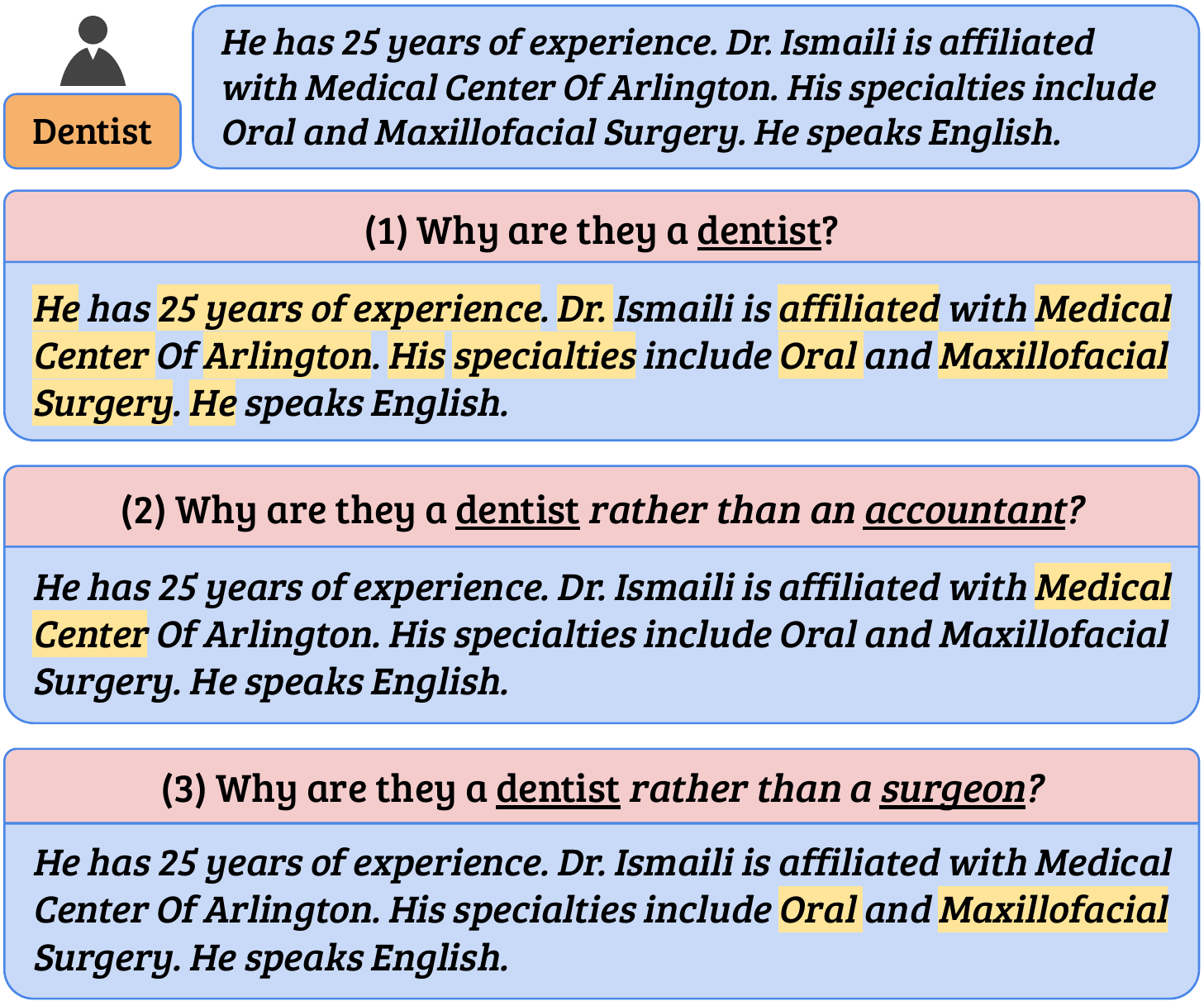}
\caption{Illustrative example of a biography from BIOS \cite{De_Arteaga_2019} with explanations for the occupation label \hl{highlighted in yellow}.
    Explanations without an explicit contrast (1) are potentially misaligned with human expectations of what is being explained, making them confusing for human interpretability.
    Contrastive explanations (2, 3) prune the space of all causal factors to `intuitively' relevant ones, aiding finer-grained understanding, and can vary based on the contrast decision (e.g. accountant, surgeon).
}
\label{fig:intro-example}
\end{figure}

\section{Introduction} 
\label{sec:intro}

Explanations in machine learning attempt to uncover the causal factors leading to a model's decisions.
Methods for producing model explanations often seek all causal factors at once---making them difficult to comprehend---or organize the factors via heuristics, such as gradient saliency \cite{simonyan2013deep,li2016understanding}. 
However, it remains unclear what makes a particular collection of causal factors a good explanation.


Studies in social science establish that human explanations, conversely, are typically ``contrastive'' \cite{miller2017social}: they rely on the causal factors that explain why an event occurred \textit{instead of} an alternative event \cite{lipton1990contrastive}.
Such explanations promote easier communication by pruning the space of all causal factors, reducing cognitive load for the explainer and explainee, as illustrated in Figure~\ref{fig:intro-example}. 
This reduction is deeply relevant: since explanations of opaque ML and NLP models are \textit{approximations} of complex statistical processes, humans interpret these decisions subjectively by assuming some contrast decision.


Our work seeks to design explanations that are explicitly contrastive, thereby revealing fine-grained aspects of model decisions, while being more representative of human comprehension (\S\ref{sec:background-contrastive}).
We introduce a novel framework for deriving contrastive explanations applicable to any neural classifier (\S\ref{sec:framework}).
Our method operates on the input representation space, and produces a latent, contrastive representation. 
We accomplish this by projecting the latent representations of the input to the space that minimally separates two decisions in the model.
We additionally propose a measure of contrastiveness (\S\ref{subsec:measure}) by computing changes to model behavior before and after the projection.

Our experiments consider two well-studied NLP classification benchmarks: MultiNLI (\citealp{multinli}; \S\ref{sec:e-snli}) and BIOS (\citealp{De_Arteaga_2019}; \S\ref{sec:bios}).
In each, we study explanations in the form of high-level \textit{concept} features or low-level \textit{textual highlights} in the input.
Our contrastive explanations uncover input features useful for and against particular decisions, and can answer which alternative label was a decision made against; this has potential implications for model debugging.
Overall, we find that a contrastive perspective aids fine-grained understanding of model behavior.


\begin{figure}[t]
\centering
\includegraphics[width=0.99\linewidth]{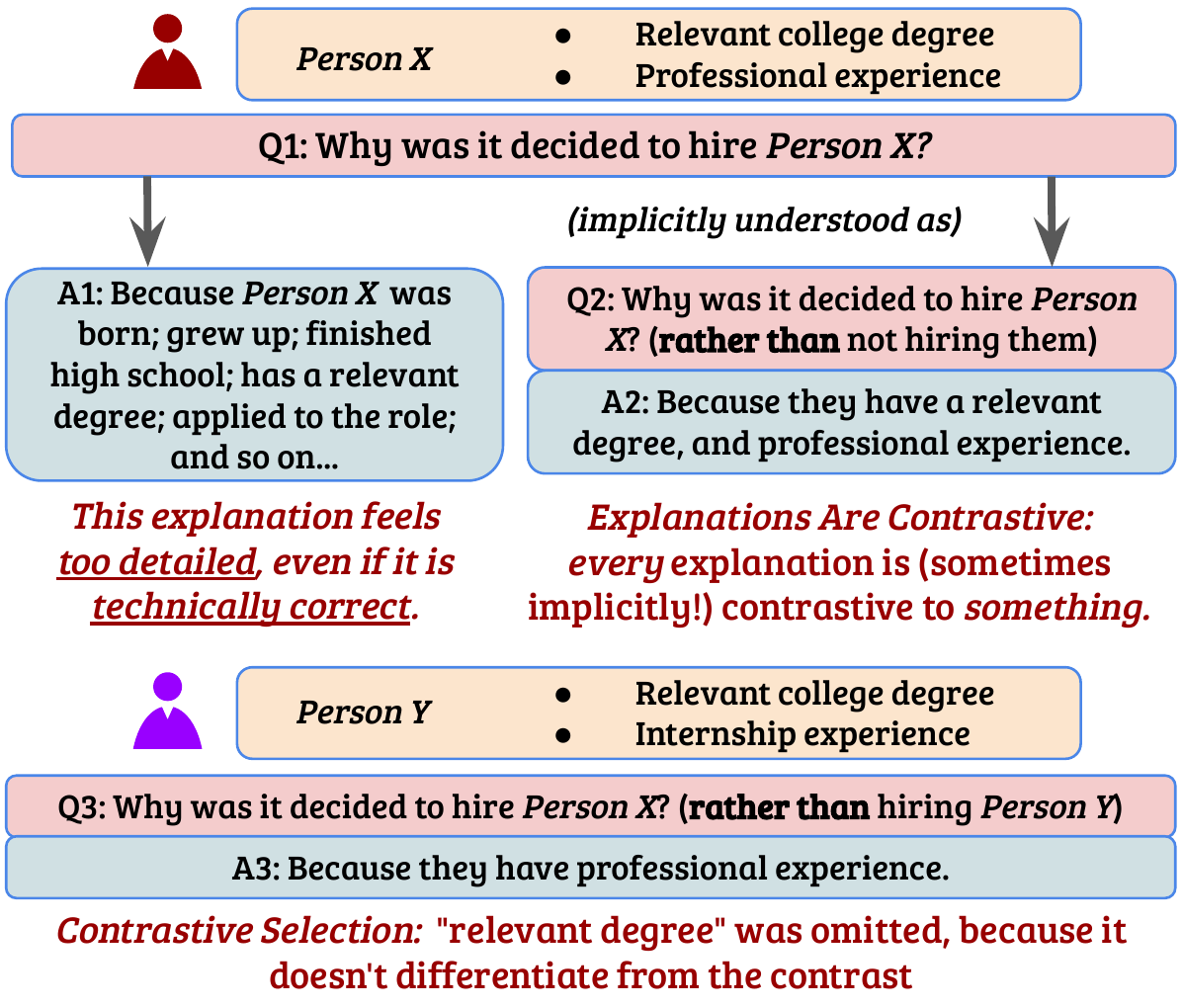}
\caption{Illustrating contrastive (A2, A3) and non-contrastive (A1) explanations (\S\ref{sec:background-contrastive}): humans intuitively produce and interpret explanations contrastively.
}
\label{fig:unwarranted-ex}
\end{figure}

\section{Contrastive Explanations}
\label{sec:background-contrastive}

Explanations can be considered as answers to the question: ``why $P$?'' where $P$, the \textbf{fact}, is the event to be explained. 
Consider the question (Q1; Fig.~\ref{fig:unwarranted-ex}):
\begin{quote}
\indent Why was it decided to hire \textit{[Person X]}?
\end{quote}
\noindent The explanation behind decision $P$ (the hiring decision) comprises the causal chain of events that led to $P$; a `reasonable' answer to the question may cite that \textit{Person X} has relevant degrees or professional experience.
However, explaining the \textit{complete} causal chain (A1; Fig.~\ref{fig:unwarranted-ex}) is both burdensome to the explainer and cognitively demanding to the explainee \cite{hilton1986knowledge,Hesslow1988}. 
For instance, the hiring event above is caused by \textit{Person X}'s application for the role---yet a reasonable explainer will likely omit this factor from the explanation for simplicity, thus reducing the cognitive load. 
But which factors should be omitted, and which should not?

The theory of contrastive explanations provides a solution common in human explanations, which inherently answer the question: ``why $P$, rather than $Q$?'' \cite{hilton1988logic}, where $Q$ (the \textbf{foil}) is some alternative event.\footnote{
Generally, $Q$ is a \textbf{contrast fact} \cite[\S4,][]{miller2018contrastive}---either a `foil' (obtained by artificially altering the input context, as in a counterfactual; see \S\ref{sec:causal-intervention}) or a `surrogate' (in a naturally-occurring context, as a bifactual). 
Our work considers only counterfactual contexts, hence we use the term `foil'.
}
Therefore, the explanation to the hiring decision might answer (Q2; Fig.~\ref{fig:unwarranted-ex}):
\begin{quote}
    Why was it decided to hire \textit{[Person X]}, rather than not hiring them?
\end{quote}
Since the decision not to hire the candidate can also be traced back to the fact that they applied to the role, the explanation can be simplified by omitting this factor (A2; Fig.~\ref{fig:unwarranted-ex}); this illustrates \textit{contrastive causal attribution} (\S\ref{ssec:contrastive-projection}) in the explanation process.
Similarly, given a different foil (Q3; Fig.~\ref{fig:unwarranted-ex}): 
\begin{quote}
    Why was it decided to hire \textit{[Person X]}, rather than hiring \textit{[Person Y]}?
\end{quote}
the explainer might find it unnecessary to include attributes (such as a ``relevant college degree'') common to both \textit{Person X} and \textit{Person Y} (A3; Fig.~\ref{fig:unwarranted-ex}). 
For the rest of the paper, we will refer to explanations which are not explicitly contrastive, such as A1 in Fig.~\ref{fig:unwarranted-ex}, as \textbf{non-contrastive} explanations.


\paragraph{Implications for model explanations:} 
 
Model explanations can benefit from explicit contrastiveness in two ways.
    Model decisions are complex and noisy statistical processes---`complete' explanations are difficult \cite{jacovi-goldberg-2020-towards}. 
    Contrastive explanations make model decisions \textit{easier to explain} by omitting many factors, given the \textit{relevant} foils.
    This reduces the burden of the explaining algorithm to interpret and communicate the `complete' reasoning. 
    Additionally, humans tend to inherently, and often implicitly, comprehend (model) explanations contrastively. 
    However, the foil perceived by the explainee may not match the one truly entailed by the explanation \cite[e.g.,][]{DBLP:conf/icml/KumarVSF20}; contrastive explanations make the foil explicit, and are therefore \textit{easier to understand}.


\section{Contrastive Explanations Framework} 
\label{sec:framework}


We present our framework for producing contrastive explanations.
We describe the preliminaries for our framework (\S\ref{subsec:structure}), and outline an interventionist approach for causal attribution (\S\ref{sec:causal-intervention}).
Next, we introduce our projection-based method to produce latent contrastive representations (\S\ref{ssec:contrastive-projection}) and a measure of the resulting behavioral change (\S\ref{subsec:measure}).

\subsection{Preliminaries}
\label{subsec:structure}

\paragraph{Candidate Factor Space:} 
All causal factors that could possibly lead to the model's decision.
These could include discrete features in the input (\textbf{textual highlights}; \citealp[]{lei16}; see Fig.~\ref{fig:intro-example}), abstract input features (\textbf{concepts}; \citealp{kim2017interpretability}; see Table~\ref{tab:bios-ex} for an example of gender as a concept), or {influential examples} in the training set for a prediction \cite{wei2017influence-functions,han-etal-2020-influence-functions-nli}.\footnote{Such spaces can also be a hybrid or partial combination of the above homogeneous spaces we consider in this work.}
Once defined, a subset of factors from the candidate factor space can then be causally attributed to the model decision.
Our framework is agnostic to the type of factor in the candidate factor space, so long as it is possible to intervene on their presence (\S\ref{sec:causal-intervention}).
Our work focuses on textual highlights and concepts.

\paragraph{Event Space:} 
The union (as a discrete set) of all possible model decision classes.\footnote{We do not consider explanations for other aspects of model behavior: such as why a particular class was assigned a particular probability, or why a particular neuron received a particular activation. Additionally, for the foil we only consider a single other class (aside from the fact), rather than some subset of classes.}
This includes the event we attempt to explain (the fact), i.e. a trained model's decision\footnote{
The fact is not strictly required to be the model prediction; it could alternatively be the model probability, for instance.
}, as well all other classes (the foils), considered independently.

\subsection{Causal Attribution via Interventions}
\label{sec:causal-intervention}

Given a candidate factor space, we seek to attribute factors with causality over the decision process, i.e. select factors which caused the model's decision.
We adopt an interventionist approach \cite{Woodward2003-WOOMTH}---this involves determining causality of a factor by intervening on it, thereby producing a \textbf{counterfactual}.
The importance of the intervened factor is determined by change in model behavior under this counterfactual. 

Most interventions we consider are\textbf{ amnesic}, i.e. use counterfactuals which \textit{omit} the candidate factors under consideration.\footnote{
An alternative is to replace the candidate factor with other causal factors. 
We leave extensive comparisons of amnesic and non-amnesic interventions to future work.
}
Factors in the form of highlights and concepts involve different kinds of interventions.
For the former, we simply replace each highlighted token with a `mask' token (\S\ref{subsec:highlight-ranking-snli},\ref{ssec:bios-pronouns}), and train models where such masked data is in distribution \cite{DBLP:conf/iclr/ZintgrafCAW17,kim2020marginalization}, i.e. pre-trained masked language models, such as \texttt{RoBERTa} \cite{roberta}.
For conceptual interventions, we employ an \textbf{amnesic operation} to remove a concept from the input representation (\S\ref{ssec:nli-concepts},\ref{subsec:bios-gender}).
Following \citet{elazar2020amnesic}, this amnesic operation uses a null-space projection to iteratively remove all linear directions correlated with the concept, until it is not possible to linearly discover the concept information from the latent vector \cite{ravfogel-etal-2020-null}.\footnote{
This method only serves to remove linear information, and is not guaranteed to remove non-linear information.
As such, an absence of behavioral change can only indicate that the final layer made no use of the inspected concept, but it may have been used in a previous layer.}
Training an amnesic probe requires labels indicating presence of the concept for each example ( App.~\ref{app:intervention_implementation}).
Where possible, we also employ a conceptual intervention via \textit{manually-annotated counterfactuals} \cite{DBLP:conf/iclr/KaushikHL20, gardner-etal-2020-evaluating} which involve textual modifications to existing data instances (\S\ref{ssec:nli-concepts}; Hyp-Negation).

\subsection{Contrastive Attribution}
\label{ssec:contrastive-projection}

While the aforementioned interventions select the causal factors from the candidate factor space for model explanations, we are specifically interested in selecting factors that yield contrastive explanations. 
Given a space of discovered causal factors, we therefore need an additional intervention to attribute contrastive behavior to a subset of these factors.
We propose a method below to produce contrastive explanations in the form of a dense representation of the input in the latent space.
This representation is given by a projection operation, such that only the components that distinguish the fact from the foil are preserved.

Formally, let $\seq{x}$ be the text input to be classified as one of $K$ output classes $\mathcal{Y} = \{y_1, ..., y_K\}$. 
Consider the model class $f$, which commonly uses an arbitrarily deep neural encoder $\text{\textit{enc}}(\cdot)$ that transforms the input $\seq{x}$ into a vector $\seq{h}_\seq{x} \in \mathbb{R}^d$.
Once encoded, a final linear layer, $\matr{W} \in \mathbb{R}^{K \times d}$ can then be applied to the input to yield the logits of the model over the $K$ classes, such that $f(\seq{x})=\matr{W}\text{\textit{enc}}(\seq{x}) = \matr{W}\seq{h}_\seq{x}$. 
Let $y^* = \argmax_{y \in \mathcal{Y}} f(\seq{x})$ be the model prediction (the \emph{fact}), $y'$ be an alternative prediction of interest (the \emph{foil}), and ${p} := \text{\textit{softmax}}(\matr{W} \seq{h}_\seq{x})$ the normalized model probabilities.


Recall that output of the model, $f(\seq{x}) = \matr{W}\seq{h}_\seq{x}$ is linear in the latent input representation, $\seq{h}_\seq{x}$.
Let ${\seq{w}_i}$ be the row in $\matr{W}$ corresponding to class $i$. 
The logits for classes $y',y^*$, given by the dot products
${\seq{w}_{y'}}^T\seq{h}_\seq{x}$ and ${\seq{w}_{y^*}}^T\seq{h}_\seq{x}$ respectively, are thus \emph{unrelated} to any other row in the prediction matrix.
These dot products are un-normalized projections of the representation $\seq{h}_\seq{x}$ on the directions ${\seq{w}_{y^*}}, {\seq{w}_{y'}} \in \mathbb{R}^d$.
While the representation $\seq{h}_\seq{x}$ can be high-dimensional, only two directions (components) in this high-dimensional space are relevant for each contrastive decision.\footnote{Two directions at the most, or one if $\seq{w}_{y^*}$ and $\seq{w}_{y'}$ share the same direction.}
Furthermore, we are interested in the prediction of one class over the other, as opposed to the logits. 
Hence, we can replace these two directions with a \emph{single} contrastive direction ${\seq{u}} \in \mathbb{R}^d$, by defining ${\seq{u}} = {\seq{w}_{y^*}} - {\seq{w}_{y'}}$. 

Given that the model favors $y^*$ over $y'$, i.e., $p_{y^*} > p_{y'}$, if and only if ${\seq{u}}^T\seq{h}_\seq{x} > 0$, the projection of $\seq{h}_\seq{x}$ onto ${\seq{u}}$ precisely yields the linear direction in $\mathbb{R}^d$ which the model uses to differentiate between classes $y^*$ and $y'$. 
We refer to the \textit{span} of the $\seq{u}$---that is, the collection of all vectors $\alpha \seq{u}$ for a scalar $\alpha$---as
the \emph{contrastive space} for $y^*$ and $y'$.
We define the contrastive transformation $C(\seq{h}_\seq{x})_{y',y^*}$ to be the orthogonal projection onto this subspace: 
\begin{align}
\label{eq:contrastive-projection}
   C(\seq{h}_\seq{x})_{y',y^*} :=& \frac{\seq{u}\seq{u}^T}{\seq{u}^T\seq{u}}\seq{h}_\seq{x}=\matr{P}_{\seq{u}}\seq{h}_\seq{x},
\end{align}
where $\matr{P}_{\seq{u}}:=\frac{\seq{u}\seq{u}^T}{\seq{u}^T\seq{u}}$ is a projection matrix onto ${\seq{u}}$.\footnote{To further motivate the use of ${\seq{u}}$, recall that all directions $\seq{v}$ orthogonal to ${\seq{u}}$ form the \emph{nullspace} of ${\seq{u}}$ and satisfy ${\seq{v}}^T{\seq{u}}={\seq{v}}^T(\seq{w}_{y^*} - \seq{w}_{y'})={0}$, or, equivalently, ${\seq{v}}^T\seq{w}_{y^*} = {\seq{v}}^T\seq{w}_{y'}$. 
Such directions support $y'$ and $y^*$ to the same extent, and are thus contrastively irrelevant; we can discard those directions from $\seq{h}_\seq{x}$, without influencing the logits for $y'$ and $y^*$. 
}
The resulting representation $C(\seq{h}_\seq{x})_{y',y^*}$ is a latent vector of the same dimensions as $\seq{h}_\seq{x}$; computing this can be understood as a contrastive intervention.
Intuitively, it captures (precisely) the latent features in $\seq{h}_\seq{x}$ which are used by the model to differentiate the fact from the foil, where $\seq{h}_\seq{x}$ is the hidden representation of $\seq{x}$ before the final classifying layer.

\subsection{Measuring Contrastive Behavior}
\label{subsec:measure}

We consider a measure of contrastive behavior, based on our interventionist approach.
Let ${q} := \text{\textit{softmax}}(\matr{W} \seq{h}_\seq{x'})$ be the model probabilities following an intervention (contrastive or otherwise) which produces counterfactual $\seq{x'}$.
Our measure of contrastive model behavior is simply the difference between normalized probabilities of the fact before and after the intervention, given by:
\begin{align} \label{eq:contrastive-measure}
    \delta_{p,q}^\text{contr} :=& \frac{p_{y^*}}{p_{y^*}+p_{y'}}-\frac{q_{y^*}}{q_{y^*}+q_{y'}}.
\end{align}
Here, the normalization ensures we consider only the fact ($y^*$) and foil ($y'$), other classes being irrelevant.
This measure can be applied to both contrastive interventions, involving $C(\seq{h}_\seq{x})_{y',y^*}$  or just causal ones, involving $\seq{h}_\seq{x}$.
Given $-1 \leq \delta_{p,q}^\text{contr} \leq 1$, our metric is constrained; the magnitude indicates the degree of contrastive behavior. 
Our metric is reminiscent of statistical parity metrics used in algorithmic fairness \cite{zemel13learning}.\footnote{
A similar metric, TRITE \cite{feder2020causalm}, evaluates causality of a factor by measuring the difference of average probabilities of all classes after a counterfactual intervention.}
We use the contrastive measure in two settings:
\begin{compactenum}
    \item \textbf{Ranking factors (controlled foil):} given a fixed foil, and the candidate factor space, we rank each factor by how contrastively useful it is to the model for choosing the fact, against the given foil. 
    \item \textbf{Ranking foils (controlled factor):} given a causal factor,     
    we rank the set of available foils in the event space by how much the said factor is contrastively used by the decision process between the fact and the foils.
\end{compactenum}
The above is a relative and continuous perspective on contrastive selection, compared to a discrete and binary one discussed in \S\ref{sec:background-contrastive}. 
We leave a possible discretization of this process to future work.

\section{Case Study \RNum{1}: Analyzing NLI}
\label{sec:e-snli}

We apply our contrastive framework to the \textit{natural language inference} (NLI) task \cite{dagan2005pascal}, on two datasets: MultiNLI \cite{multinli} and SNLI \cite{bowman2015large}.
Given a \textit{premise} sentence, the NLI task classifies if a \textit{hypothesis} sentence \textit{ entails}, \textit{contradicts} or is \textit{neutral} to the premise. 
Our experiments are based on a \verb+RoBERTa+-large model \cite{roberta} fine-tuned on MultiNLI (obtaining 90.1\% accuracy on dev-mismatched), unless otherwise specified; see more details in Appendix~\ref{app:model_implementation}.

\paragraph{Sanity Checks.}
Our first set of experiments use a controlled setup to verify that our method works as expected.
We train a MultiNLI model on modified instances with label-specific ``stains'' (input tokens inserted to contrastively distinguish classes; \citealp{datastaining}).
We intervene on the stains (as highlights; \S\ref{sec:causal-intervention}) before our contrastive projection (\S\ref{ssec:contrastive-projection}), and then rank highlights and foils by \metric (\S\ref{subsec:measure}).
The most contrastive foils and highlights indeed correspond to the stains, verifying our methodology; see results and details in App.~\ref{app:staining}.

\begin{table}[t]
\centering
\resizebox{0.98\linewidth}{!}{%
\begin{tabular}{lrrrrrrr} \toprule
            \bf \multirow{2}{*}{\bf Concept} & \multirow{2}{*}{\bf Tot.} & \multicolumn{3}{c}{\bf Gold} & \multicolumn{3}{c}{\bf Predicted}\\
            \cmidrule(lr){3-5} \cmidrule(lr){6-8}
            & & \%E & \%C & \%N & \%E &\%C & \%N \\ \midrule
Overlap     &  5.7 & 63.7 & 29.6& 6.7&  64.4&  29.2& 6.3  \\
Hypothesis  & 52.3 & 33.9 & 33.1& 32.9&  49.1& 24.7& 26.2  \\
Hyp-Neg     & 14.8 & 21.4 & 61.0&  17.6&  21.6&  60.7&  17.7  \\
\bottomrule
\end{tabular}
}
\caption{Prevalence of overlap, hypothesis, and hyp-negation concepts in MultiNLI dev (shown as \%).
\textbf{Gold} indicates \% examples with the concept by gold labels, \textbf{Predicted} with respect to a \textit{RoBERTA} model predictions, and \textbf{Tot.} shows dataset aggregates.
E, C, N stand for entailment, contradiction and neutral, resp. 
}
\label{tab:esnli-concept-distribution} 
\end{table}

\subsection{The Role of NLI Concepts} 
\label{ssec:nli-concepts}

Extensive prior work supports the presence of spurious correlations or artifacts in popular NLI datasets \cite{poliak-etal-2018-hypothesis,DBLP:conf/naacl/GururanganSLSBS18}.
For e.g., instances with high lexical overlap between premise and hypothesis tend to correlate with the \textit{entailment} class, and those containing negations in the hypothesis with the \textit{contradiction} class.
As a result, models trained on these datasets tend to rely on these features to make accurate predictions, \textit{regardless} of other semantic signals.
Moreover, \citet{DBLP:conf/naacl/GururanganSLSBS18} show that a model can ignore the premise altogether and still make an accurate prediction based only on the hypothesis.
Table~\ref{tab:esnli-concept-distribution} shows the distribution of the above three concepts in  MultiNLI.
While prior work provides considerable evidence that NLI model decisions rely heavily on these concepts, we investigate whether this reliance is \textit{contrastive} by nature. 
We consider each concept independently:

\begin{table}[t]
\centering
\resizebox{0.98\linewidth}{!}{%
\begin{tabular}{llrrr} \toprule
\multirow{2}{*}{\bf concept} & \multirow{2}{*}{\bf fact} & \multicolumn{3}{c}{\bf foil} \\
\cmidrule(lr){3-5}
            &   & E & C & N  \\ \midrule
Overlap & E                     & - & 0.006    &   0.433   \\ 
Hypothesis (MultiNLI)  & E      & - & -0.005     & -0.031  \\
Hyp-Negation & C                & 0.195 & - & 0.051 \\\bottomrule
\end{tabular}
}
\caption{ 
Contrastive effects of NLI concepts \S\ref{ssec:nli-concepts} on MultiNLI dev as given by $\delta_{p,q}$ following causal and contrastive interventions. 
E, C, and N indicate entailment, contradiction, and neutral, resp.
}
\label{tab:esnli-concept-amnesic-probing}
\end{table}

\paragraph{Overlap.} Instances with the overlap concept are those where all of the content words\footnote{Based on \textit{spaCy}'s list of English stop-words.} in the hypothesis also exist in the premise (in any order).
Prior work has shown that the overlap concept is highly relevant in the model's reasoning process for predicting entailment \cite{naik-etal-2018-stress,mccoy-etal-2019-right}. 
We intervene on the overlap concept via amnesic probing (\S\ref{sec:causal-intervention}), followed by our contrastive intervention (\S\ref{ssec:contrastive-projection}) to measure behavioral changes (\S\ref{subsec:measure}).
Foil ranking results in Table~\ref{tab:esnli-concept-amnesic-probing} show that when predicting \textit{entailment}, the \textit{overlap} concept is overwhelmingly contrastive against \textit{neutral}. 
This aligns with Table~\ref{tab:esnli-concept-distribution} statistics, which show that the concept is highly correlated with \textit{entailment} (64.4\%) and against \textit{neutral} (6.3\%)  predictions. 
The overlap concept is thus \textit{contrastively important}.

\paragraph{Hypothesis.} 
Motivated by the finding that hypothesis-only models have been shown to achieve high accuracy in the NLI task \cite{DBLP:conf/naacl/GururanganSLSBS18,poliak-etal-2018-hypothesis}, we consider the `hypothesis' concept---a collection of all concepts existing only in the hypothesis.
This concept is realized in instances accurately predicted by a hypothesis-only baseline.
We use binary concept labels based on accurate / inaccurate predictions of a hypothesis-only \verb+RoBERTa+-baseline to train an amnesic probe for causal intervention (\S\ref{sec:causal-intervention}).
The amnesic probe and our contrastive intervention (\S\ref{ssec:contrastive-projection}) are then applied on the full-input model.
Results in Table~\ref{tab:esnli-concept-amnesic-probing} show that when predicting \textit{entailment} in MultiNLI, this concept is \textit{not} strongly contrastive to either foil (-0.005 v. -0.031). 
This aligns with Table~\ref{tab:esnli-concept-distribution} statistics, since the concept is similarly distributed with \textit{contradiction} and \textit{neutral} (24.7\% v. 26.2\%). 
However, when applied to the SNLI dataset, we see stronger contrastive behavior with respect to \textit{contradiction} (0.505) than \textit{neutral} (0.463).
Perhaps this could be explained by the higher hypothesis-only bias in SNLI, compared to MultiNLI \cite{DBLP:conf/naacl/GururanganSLSBS18}.


\paragraph{Hyp-Negation.} This concept is realized in instances containing negation words (e.g., `no') in the hypothesis.
The presence of this concept is highly indicative of an NLI model's prediction to be the \textit{contradiction} class regardless of other NLI semantics \cite{DBLP:conf/naacl/GururanganSLSBS18}. 

Here, we use manually annotated counterfactuals to intervene on the concept\footnote{
Initial experiments with amnesic probing for this concept were inconclusive. 
We suspect that although useful for the model, the concept is perhaps not detectable in the last layer---the only conclusion the amnesic probing method can draw. 
}.
Given an example without negations in the hypothesis, and predicted by the model as \textit{entailment} or \textit{neutral}, two of the authors manually paraphrased the hypothesis to include a negation \textit{without altering the semantics}; see Appendix~\ref{app:negation-concept-ex} for examples.\footnote{Our collection of 90 such instances for each of \textit{entailment} and \textit{neutral} will be released upon publication.} 
We then proceed to probe the model for behavioral changes ($\delta_{p,q}^\text{contr}$) between the instances before--and--after the intervention, treating the negated example as a counterfactual to compute $q$. 
Foil ranking results in Table~\ref{tab:esnli-concept-amnesic-probing} show that, on average, the model utilizes the negation concept as evidence for \textit{contradiction} in contrast to \textit{entailment} (0.195), as opposed to \textit{neutral} (0.051). 

 
In summary, while it is known that the above three concepts in NLI are useful, we investigated whether they are useful \textit{against a particular foil}. 
Two concepts (overlap and hyp-negation) do have a prominent contrastive role, while the hypothesis concept is contrastive in only one setting, indicating that there might be concepts which are not explicitly contrastive.

\begin{table}[t]
\centering
\small
\resizebox{0.98\linewidth}{!}{%
  \begin{tabular}{llp{6.5cm}} 
  \toprule
Fact             & Foil (\textbf{gold})      & Input with Highlights \\ \midrule
\multirow{5}{*}{\rotatebox[]{90}{{entailment}}} &  & P: A nun uses her camera to take a photo of an interesting site.      \\
& none    & H: A nun taking photos of a \hl{interesting} site outside. \\
& contradict. & H: A \hl{nun} taking photos of a interesting site outside.     \\
& \textbf{neutral}    & H: A nun taking photos of a \hl{interesting} site outside.   \\  \midrule
\multirow{8}{*}{\rotatebox[]{90}{neutral}} &  & P: A couple bows their head as a man in a decorative robe reads from a scroll in Asia with a black late model station wagon in the background.     \\
& none   & H: A \hl{light} black late model station wagon is in the background. \\
& \textbf{entailment} & H: A \hl{light} black late model station wagon is in the background.    \\
& contradict.    & H: A light \hl{black} late model station wagon is in the background.   \\  \midrule
\multirow{4}{*}{\rotatebox[]{90}{neutral}} &  & P: Girl plays with colorful letters on the floor.    \\
& none   & H: The girl is having fun \hl{learning} her letters. \\
& \textbf{entailment} & H: H: The girl is having fun \hl{learning} her letters.    \\
& contradict.    & H: The girl is having \hl{fun} learning her letters.   \\  \midrule
\multirow{8}{*}{\rotatebox[]{90}{neutral}} &  & P: Three men with blue jerseys try to score a goal in soccer against the other team in white jerseys and their goalie in green.     \\
& none   & H: Some men with jerseys are in a \hl{bar}, watching a soccer match. \\
& entailment & H: Some men with jerseys are in a \hl{bar}, watching a soccer match.    \\
& \textbf{contradict.}    & H: Some men with jerseys are in a bar, watching a \hl{soccer} match.
         \\   \bottomrule
\end{tabular} 
}
\caption{Contrastive explanations for interpreting model errors via \hl{highlight} ranking.
The fact is the incorrect prediction, and one of the foils is the gold label.
The \textit{`none'} foil indicates non-contrastive explanations (highlight intervention only).
One of the contrastive explanations agrees with the non-contrastive one.
}
\label{tab:esnli-highlights}
\end{table}

\subsection{Ranking Highlights for Debugging NLI}
\label{subsec:highlight-ranking-snli}

Contrastive explanations can help humans understand model errors.
Our goal is to answer: what factor led to the model's incorrect prediction \textit{in contrast to} the gold label?
We achieve this by treating an erroneous model prediction as the fact, and the gold label as the foil.
For this fact-foil pair, we rank different factors based on their contrastive behavior (\S\ref{subsec:measure}).\footnote{We report a BIOS factor ranking experiment in App.~\ref{app:highlight-ranking-bios}.}
We consider all unigrams and bigrams in the hypothesis as highlight factors in the NLI task. 
As before, we intervene on each factor, apply our contrastive intervention, and measure the change in behavior (\S\ref{subsec:measure}).

Table~\ref{tab:esnli-highlights} presents some SNLI examples \swabha{did we use the multinli roberta though?specify!}where we report the most contrastive highlight factor (by \metric) for each foil. 
We also report a non-contrastive explanation (\S\ref{sec:background-contrastive}) resulting from only the causal (highlight) intervention (i.e. no contrastive projection).
We see that one of the contrastive explanations (usually with the gold foil) agrees with the non-contrastive explanation, indicating that the latter might simply be reflecting an implicit contrastive explanation.
The last row shows a case where the non-contrastive explanation does not agree with the contrastive explanation for the gold foil.
However, the model's reasoning appears to be correct since the hypothesis may or may not entail the premise; additionally `bar' seems the correct reasoning for choosing \textit{neutral} over \textit{entailment}.
Thus, contrastive explanations can provide insight on why the model specifically preferred its prediction \textit{over the gold label}. 
Future work might explore using contrastive explanations to detect labeling errors in NLI \cite{swayamdipta-etal-2020-dataset}.



\section{Case Study \RNum{2}: Analyzing BIOS} 
\label{sec:bios}

\begin{table}[t]
\centering
\small
\resizebox{\linewidth}{!}{   %
   \begin{tabular}{p{7.7cm}} 
        \toprule \bf Biography / Profession / Gender \\ \midrule
        \hl{She} also works as a Restitution Specialist while being the liaison to the Victim Compensation Board. \hl{Ms. Azevedo} was named an OVSRS Outstanding Partner due to \hl{her} dedication to providing critical information to staff so victims can obtain their court-ordered restitution while offenders can be held accountable. 
        / \textbf{paralegal} / \textbf{F} \\ \midrule
        \hl{Peter} also has substantial experience representing clients in government investigations, including criminal and regulatory investigations, and internal investigations conducted on behalf of clients. / \textbf{attorney} / \textbf{M}  \\ \bottomrule
\end{tabular}
}
\caption{
Examples from the BIOS train-set.
The \hl{highlights in yellow} indicate demographic information in individual biographies, encoded as names and pronouns.
These highlights could be factors explaining a model's prediction of the associated profession.
}
\label{tab:bios-ex}
\end{table}

We apply our framework on the BIOS dataset \cite{De_Arteaga_2019} containing individuals biographies, labeled with their professions and binary gender\footnote{
We acknowledge that this is a simplification and erases those who do not identify with this binary.} (see Table~\ref{tab:bios-ex}).
The task involves classifying a biography as one of 27 professions\footnote{We omit the 28th profession (\textit{model}) from the data, as we found the annotations inconsistent; see App.~\ref{app:model_implementation}.}, \textit{without} explicitly considering the gender attribute.
We analyze \verb+RoBERTa+-large \cite{roberta} fine-tuned on BIOS, with test performance of 87.52\%.

\begin{table}[t]
\centering
\resizebox{0.95\linewidth}{!}{%
  \begin{tabular}{llcc}
        \toprule 
        \textbf{fact} (\% males) & \textbf{foil} (\% males) &  \metric & \bf sign(\textit{cos}) \\ 
        \midrule
        paralegal (9\%) & attorney (62\%) & 0.804 & $-$ \\
        professor (55\%) & teacher (41\%) & 0.225 & $+$ \\
        accountant (63\%) & psychologist (37\%) & 0.108 & $+$ \\
        nurse (9\%) & physician (47\%) & 0.084 & $-$ \\
        teacher (41\%) & poet (53\%) & 0.072 & $-$ \\
        \bottomrule
\end{tabular}}
\caption{ The top-5 fact-foil pairs in the BIOS dataset for which the gender concept was measured to have the largest contrastive effect; the most contrastive foil is shown for each fact. 
The ``male'' concept is supportive of \textit{attorney} over \textit{paralegal}, reflective of BIOS gender statistics, where \textit{attorney} is a male majority profession.
}
\label{tab:bios-concept-results}
\end{table}

\begin{table}[t]
\ra{1.15}
\centering
\small
\resizebox{0.9\linewidth}{!}{%
\begin{tabular}{llclc} 
\toprule
\multirow{2}{*}{\textbf{fact ($y^*$)}}  & \multicolumn{2}{c}{\textbf{Most Contrastive}} & \multicolumn{2}{c}{\textbf{Least Contrastive}} \\
\cmidrule(lr){2-3} \cmidrule(lr){4-5}
     & \textbf{foil}  & $\delta_{p,q}^\text{contr}$\% & \textbf{foil}   & $\delta_{p,q}^\text{contr}$\%   \\ 
\midrule
 \multirow{3}{*}{paralegal}   &  attorney & \hlmost{10.519} &  composer & \hlleast{0.019}  \\
 &  accountant & \hlmost{1.165} &  dj & \hlleast{0.021} \\ 
 &  professor & \hlmost{0.387} &  surgeon & \hlleast{0.026} \\ \midrule
\multirow{5}{*}{physician}  &   professor   &  \hlmost{0.622}       &    rapper & \hlleast 0.005 \\         
  &   surgeon   &    \hlmost{0.406}      &   composer & \hlleast 0.006  \\       
  &   psychologist   &   \hlmost{0.185}   &      dj & \hlleast 0.006  \\     
  &   $\dagger$ nurse     &   \hlneg{-0.315}    & - & -  \\ 
  &   $\dagger$ chiropractor    &  \hlneg{-0.132} & - & -  \\ 
  \midrule
\multirow{4}{*}{attorney}  &  professor & \hlmost{1.292} & composer & \hlleast{0.018} \\
 &  journalist & \hlmost{0.545}  & rapper & \hlleast{0.022} \\
 &  teacher & \hlmost{0.420} &  comedian & \hlleast 0.023  \\   
 & $\dagger$ paralegal & \hlneg{-0.223} & - & - \\ \midrule
\multirow{3}{*}{nurse}   &  physician & \hlmost{3.531} &  paralegal & \hlleast{0.067} \\
 &  surgeon & \hlmost{1.779} &  composer & \hlleast{0.105} \\
 &  chiropractor & \hlmost{1.762} & interior designer & \hlleast{0.117}  \\ \midrule
 \multirow{3}{*}{rapper}  &  dj & \hlmost{1.878} &  paralegal & \hlleast{0.130} \\
 &  composer & \hlmost{1.844} &  dietitian & \hlleast{0.177} \\
 &  poet & \hlmost{1.335} &  interior designer & \hlleast{0.272} \\ \midrule
  \multirow{3}{1cm}{interior designer}  &  architect & \hlmost{4.095} &  composer & \hlleast{0.081} \\
 &  photographer & \hlmost{2.946} &  chiropractor & \hlleast{0.093} \\
 &  journalist & \hlmost{2.046} &  paralegal & \hlleast{0.105} \\
 \bottomrule
\end{tabular} }
\caption{
The most and least contrastive foils by \metric\% for five profession predictions (facts), aggregated by fact labels across BIOS dev, when intervening on the pronoun and name highlights. 
$\dagger$ indicates foils with the \sethlcolor{Lavender}\hl{largest negative \metric}; here the highlights provided strong evidence towards the foil, rather than the fact.
}
\label{tab:bios-highlights-results}
\end{table}

\subsection{The Role of the Gender Concept} 
\label{subsec:bios-gender}
\citet{ravfogel-etal-2020-null} showed that a (binary) gender concept is valuable for BIOS model predictions.
We use amnesic probing (\S\ref{sec:causal-intervention}) to intervene on the presence of the gender concept, trained using the binary gender labels in BIOS, followed by our contrastive projection (\S\ref{ssec:contrastive-projection}).

Table~\ref{tab:bios-concept-results} reports the top-5 fact/foil pairs based on the contrastive measure $\delta_{p,q}^\text{contr}$, among all pairs of professions, along with the respective binary gender \% in BIOS. 
The top-scoring pairs tend to be semantically similar while dissimilar in their gender proportions (e.g. \textit{paralegal} and \textit{attorney}).
This confirms that the model is indeed leveraging the gender concept to differentiate between otherwise semantically-similar classes. 

Amnesic probing for a concept results in a representation that cannot distinguish whether it was present or not in an instance.
But it results in a final ``concept vector", $\seq{r}$, which only maintains the information relevant towards the concept. 
In Table~\ref{tab:bios-concept-results}, we additionally report the sign of the cosine similarity between $\seq{u}$ and $\seq{r}$,
where $\seq{u} = \seq{w_{y^*}} - \seq{w_{y'}}$ (see \S\ref{ssec:contrastive-projection}).
This indicates whether the concept (in this case, ``male'') is present in the fact ($+$) or not ($-$).
The results align with intuition: the ``male'' concept is supportive of \textit{attorney} over \textit{paralegal}, and \textit{accountant} over \textit{psychologist}, which are male-majority professions in the BIOS dataset.

\subsection{The Role of Demographic Highlights} 
\label{ssec:bios-pronouns}

The demographic attributes of individuals, as encoded by their pronouns and personal names, can be spuriously correlated with their professions, as is often manifest in the BIOS dataset \cite{De_Arteaga_2019,DBLP:conf/naacl/RomanovDWCBCGKR19}.
For instance, paralegals in BIOS are overwhelmingly women (roughly 90\%); female names and pronouns might be very predictive of this profession, albeit for incorrect reasons. 
Further, names might reveal other demographics (e.g., \textit{Azevedo} is a common Portuguese surname) potentially predictive of certain professions.
Table~\ref{tab:bios-ex} shows pronouns and person name highlights which are candidate causal factors of interest. 
We investigate the contrastive importance of these factors, by asking: which classes does the model use the pronouns and names \textit{contrastively against} when making its decision?

We intervene on pronoun and name highlights by masking (\S\ref{sec:causal-intervention}), followed by computing the contrastive measure (Eq.~\ref{eq:contrastive-measure}) for every possible profession (foil) in contrast to the model prediction (fact).
Table~\ref{tab:bios-highlights-results} shows the most and least contrastive foils for five professions (facts), where the foils are ranked by \metric, and aggregated across BIOS dev.
For example, on \textit{paralegal} predictions, \textit{attorney} is the most relevant foil, indicating that the model uses demographic information for that distinction.
This indicates that the model leverages demographic attributes as evidence for decisions between classes, which are semantically similar but demographically different.
Unlike \metric values obtained from amnesic probes, the sign of \metric obtained after highlight interventions can be meaningful.
When \metric$> 0$, the model uses highlights as evidence \textit{for} the fact and \textit{against} the foil; negative values indicate evidence for the foil against the fact.
For \textit{attorney} predictions, \textit{paralegal} is an important foil as expected, even though \metric$< 0$.



\subsection{Contrastive-Only Interventions}
\label{ssec:measuring-cont-only}

We are additionally interested in measuring the degree of contrastive behavior change without considering causal features such as highlights / concepts.
We can treat the \textit{contrastive projection} $C(\seq{h}_\seq{x})_{y',y^*}$ (Eq.~\ref{eq:contrastive-projection}) as an intervention\footnote{This is an amnesic intervention (at the last layer of the model's reasoning) since it \textit{forgets} the information that cannot differentiate the fact and foil.}, and measure the change in behavior following just this intervention.
Since the contrastive intervention, by construction, precisely maintains contrastive behavior, $\delta_{p,q}^\text{contr}$ is no longer appropriate.
We thus use a symmetrized Kullback-Leibler divergence, $D_{\text{KL}}(p \; \Vert \; q) + D_{\text{KL}}(q \; \Vert \; p)$, which gives us the global behavior change across the dataset after the contrastive intervention.

\begin{table}[t]
\ra{1.15}
\centering
\small
\resizebox{0.98\linewidth}{!}{%
\begin{tabular}{llclc} 
\toprule
\multirow{2}{*}{\textbf{fact ($y^*$)}}  & \multicolumn{2}{c}{\textbf{Least Contrastive}} & \multicolumn{2}{c}{\textbf{Most Contrastive}} \\
\cmidrule(lr){2-3} \cmidrule(lr){4-5}
     & \textbf{foil}  & $D_\text{KL}$ & \textbf{foil}   & $D_\text{KL}$   \\
\midrule
\multirow{3}{*}{paralegal}   & attorney & 4.680 & surgeon & 0.779 \\
 & accountant & 2.295 & professor & 0.889 \\
 & interior designer & 1.978 & physician & 0.993 \\ \midrule
\multirow{3}{*}{physician}  &  surgeon & 3.867 & paralegal & 0.847 \\
 & professor & 2.400 & dj & 0.849 \\
 & nurse & 2.027 & photographer & 0.875  \\     \midrule
\multirow{3}{*}{attorney}   & professor & 2.581 & dj & 0.528 \\
 & paralegal & 2.195 & personal trainer & 0.576 \\
 & journalist & 1.420 & chiropractor & 0.586   \\     \midrule
\multirow{3}{*}{nurse}   & professor & 2.386 & dj & 0.740 \\
 & physician & 2.305 & software engineer & 0.747 \\
 & psychologist & 1.662 & rapper & 0.763  \\ \midrule
 \multirow{3}{*}{rapper}   & dj & 2.892 & dietitian & 0.725 \\
 & poet & 2.705 & yoga teacher & 0.942 \\
 & comedian & 1.931 & architect & 0.964 \\ \midrule
  \multirow{3}{1cm}{interior designer} & architect & 3.540 & composer & 0.869 \\
 & photographer & 2.145 & chiropractor & 1.021 \\
 & journalist & 2.054 & pastor & 1.150 \\
 \bottomrule
\end{tabular} }
\caption{Results on the measurement of contrastive power in the decision process of the model's last layer for making \textit{fact} predictions. This answers how much, on average, the model relies on differentiating  (e.g.,) \textit{physician} from \textit{surgeon} when making \textit{physician} predictions. The measurement is \textit{inverse} to the degree that the differentiation is dominating the decision process.
}
\label{tab:bios-contrastive-power-results}
\end{table}

Table~\ref{tab:bios-contrastive-power-results} reports results on the $D_{\text{KL}}$ metric for BIOS, applying a similar methodology as Table~\ref{tab:bios-highlights-results}, but without the highlights intervention. 
When predicting the fact, the contrast between the fact and a highly impactful foil does \textit{not} significantly impact the decision (since removing the non-contrastive information greatly affects the decision), and vice-versa. 
For e.g., the contrast between the related professions of \textit{attorney} and \textit{paralegal} does not substantially affect the decision to predict \textit{attorney} (2.195).
As expected, the trend is reversed for \textit{attorney} and \textit{dj} (0.528), two distant professions.

\subsection{Highlight Ranking in BIOS} 
\label{subsec:highlight-ranking-bios}


Analogous to the MultiNLI highlight ranking procedure and results presented in Section~\ref{subsec:highlight-ranking-snli}, we present highlight ranking for the BIOS task.
Here, our candidate factor space considers all word unigram and bigram highlights, for simplicity. 
We derive the model decision after intervening on each candidate, and measure the change in behavior.

We apply this technique towards understanding model errors, by selecting examples of model mistakes and assigning the foil to be the gold label. 
Qualitative examples in Table~\ref{tab:bios-ranking-factors-ex2} (Appendix \ref{app:highlight-ranking-bios}) show the top-ranking highlight for answering the question: {which unigram or bigram was most relevant for the model in making its prediction rather than the gold label?}; see Table caption for a detailed discussion.


\section{Related Work}
\label{sec:related}

The interventionist approach to causality in our work follows several recent works in NLP \citep{giulianelli2018under,meyes2018, vig2020, elazar2020amnesic, feder2020causalm}, and is justified by accumulating empirical evidence for the inability to draw causal interpretation from statistical associations alone \citep{hewitt2019,tamkin2020,ravichander2020,elazar2020amnesic}.  
Our contrastive interventions follow an amnesic operation, similar to \citet{feder2020causalm} who assess the causality of concepts, by adversarial removal of information guided by causal graphs. 
While we share the amnesic method, we focus on contrastive explanation, while they focused on the influence of concepts on model performance. 


Contrastive explanations are a relatively new area in NLP.
Recently, \citet{jacovi2020aligning} proposed to derive highlights containing the portion of the input which flips the model decision; others propose similar flips via minimal edits \cite{ross2020explaining} and conditional generation \cite{wu2021polyjuice}. 
These can be viewed as other interventions orthogonal to our work, since our contrastive framework can be used to understand such interventions.
Additionally of interest are adversarial perturbations \cite{Ganin_2017}, which are usually implemented as gradient-based interventions.
In contrast, our work relies on the identification of erasure---using linear algebra---of linear subspaces that are associated with a given concept. 
Subspace-based interventions have the advantage of being more interpretable and controlled when compared with gradient-based interventions, which, albeit expressive, are quite opaque, not to mention of unclear efficacy \cite{elazar-goldberg-2019-wheres}.

\citet{DBLP:journals/corr/abs-1906-09293} propose a model-agnostic contrastive explanation scheme based on Shapley values. They offer a local explanation, unlike our global method. 
In addition, our approach employs behavioral interventions, while \citet{DBLP:journals/corr/abs-1906-09293} do not.
Others have raised concerns regarding feature importance methods based on Shapley-values \cite{DBLP:conf/icml/KumarVSF20}; the implicit foil of such methods can be unintuitive to human explainees.

In computer-vision, many have studied the generation of counterfactual explanations or counterfactual data points.
\citet{freiesleben2020intriguing} provide a unifying theoretical framework around the relationship between adversarial examples and counterfactual explanations.  
\citet{DBLP:journals/corr/abs-1806-09809} proposed a method that provides natural language counterfactual explanation of image classification decisions. 
They have relied on a model that proposes potential counterfactual evidence, followed by a verifier that is based on human-provided image description. As their method relies on pre-existing explanation model and human descriptions, there is no guarantee the explanation it provides are related to the model's reasoning process. 
\citet{DBLP:journals/corr/abs-2004-06524} used GANs to generate examples representing minority groups, to improve fairness measures. 
This work, like other works in vision, relies on the continuous input, which is not present in natural-language applications. For more information, see \citet{9321372} for a survey of counterfactual explanations. 





\section{Conclusion}
\label{sec:conclusion}

We introduce a novel framework for producing contrastive explanations for model decisions, via a projection of the input representation to a contrastive space for the prediction and an alternate label.
We also propose a measure of the degree of contrastive behavior, following a contrastive intervention. 
Our experiments with English text classification benchmarks on BIOS and NLI demonstrate our framework's ability to rank model decisions, as well as features responsible for the decision, \textit{contrastively}.
Our quantitative and qualitative evaluations show the fine granularity of contrastive explanations, which could be useful for debugging model behavior.
Our framework is general enough to extend other (interventionist) explanation methods to produce contrastive explanations.

Contrastive explanations in NLP and ML are relatively novel; future research could explore variations of interventions and evaluation metrics for the same.
This paper presented a formulation designed for contrastive relationships between two specific classes; future work that contrasts the fact with a combination of foils could explore a formulation involving a projection into the subspace containing features from all other classes. 

\section*{Acknowledgements}
We thank the anonymous reviewers for their helpful feedback, as well as colleagues from the Allen Institute for AI.
This project has received funding from the European Research Council (ERC) under the European Union's Horizon 2020 research and innovation programme, grant agreement No. 802774 (iEXTRACT).

\bibliography{anthology,custom}
\bibliographystyle{acl_natbib}

\newpage
\clearpage
\newpage

\appendix
\label{sec:appendix}

\section{Implementation Details}
\label{app:implementation}

\subsection{Interventions}
\label{app:intervention_implementation}

We utilized three interventions in this work: masking of highlights, amnesic probing of concepts, and erasure of non-contrastive information. 

\paragraph{Masking.} The masking intervention involved replacing each token in the highlight with a predefined mask token. As we used a pre-trained masked language model for our initialization, we have used that model's mask token, which is \verb+<mask>+ in the case of \verb+RoBERTa-Large+.

\paragraph{Amnesic probing.} We have used the publicly available implementation provided by \citet{elazar2020amnesic}, which originally proposed the algorithm. Specifically, we train an \textit{iterative nullspace projection} probe until convergence which captures the linear directions which correlate with the given concept, and then project the model's latent representation on the null-space of this probe. Please refer to \citet{elazar2020amnesic} for more details.\footnote{Code available at \url{https://github.com/yanaiela/amnesic_probing}.} 

\paragraph{Contrastive projection.} In Section~\ref{ssec:measuring-cont-only} we propose to use contrastive projection as a standalone intervention to probe for the magnitude of contrastive reasoning process in the model. As mentioned in the main text, this is simply $\text{\textit{softmax}}(\matr{W}C(\seq{h}_\seq{x})_{y',y^*})$ where the original model output is $\text{\textit{softmax}}(\matr{W}\seq{h}_\seq{x})$.

\begin{table*}[t]
\centering
\small
\resizebox{\linewidth}{!}{   %
   \begin{tabular}{p{17cm}}
        \toprule \bf Biography \\ \midrule
        Shahid Kapoor is a model vegetarian. In 2011 he had been voted as Asia’s sexiest vegetarian by PETAAsiaPacific.com. After having an average year 2010, with films like Paathshala, Milenge Milenge and Badmash Company doing mediocre business, he is looking forward to the film Mausam in 2011. The film is being directed by Pankaj Kapoor and produced by Sheetal Talwar and Sunil Lulla. Sonam Kapoor and Anupam Kher acted opposite him. It is slated to be released in September, 2011. \\ \midrule
``Allison Smith is a model of tenacity and perseverance. She has battled several serious illnesses, undergone multiple surgeries, and endured life-changing procedures, yet embraces her life with joyful exuberance and optimism. Allison inspires all those who have the pleasure of meeting her and hear her incredible story. Her book has the capability of changing the lives of her readers – both those who endure chronic illnesses, as well as the caretakers (families and friends) who walk the journeys with them.'' -Karen'
\\ \midrule
Justin Bieber is a role model to the people who enjoy his music. The vast majority of these people are children, 8-16 year old girls, therefore, for him to smoke marijuana is a horrible example for the children that look up to him. It's the same as when a child sees their big sibling smoking a cigarette and feels the compulsive need to be like them.\\ \bottomrule
\end{tabular}
}
\caption{
Examples wrongly labeled with the `[fashion] model' class from the BIOS train-set.
}
\label{tab:bios-ex-model}
\end{table*}
\subsection{Models and Training}
\label{app:model_implementation}

Our experiments were implemented in AllenNLP \cite{gardner-etal-2018-allennlp} version \verb+1.2.0rc1+. 
The models used were fine-tuned \verb+RoBERTa-Large+ on the BIOS and MultiNLI training sets, and the models with the best dev-set performance among twenty epochs were chosen for analysis. 
The models otherwise used default training configurations of AllenNLP, and we provide these configurations in the repository to be included \swabha{is this done now?} \alon{yes} with this work.

In all of our training and analysis experiments, we did not use any instances labeled with the `model' class in BIOS, in all of the training, dev and test sets, due to an observation on noisy labeling made by \citet{ravfogel-etal-2020-null} which we verified. 
Table~\ref{tab:bios-ex-model} contains examples of errors in the labeling for this class. 
This lowers the number of classes in the original BIOS task from 28 to 27.

\begin{table*}[t]
\centering
\resizebox{0.98\linewidth}{!}{%
  \begin{tabular}{lllp{15cm}c} \toprule
Stained Class                          & Prediction (Fact)              & Foil       & Text \& Highlight   & Passes? \\ \midrule
\multirow{2}{*}{entailment} & \multirow{2}{*}{contradiction} &  & P: Ramses II did not build it from stone but had it hewn into the cliffs of the Nile valley at a spot that stands only 7 km (4 miles) from the Sudan border, in the ancient land of Nubia. &     \\
 &  & entailment & H: \hl{\textit{Though,}} Ramses II ordered that it be made out of stone and not hewn into the cliffs. & \cmark     \\
                               &                                & neutral    & H: \textit{Though,} Ramses II ordered that it be made out of stone and \hl{not} hewn into the cliffs. & \cmark    \\ \midrule
\multirow{1}{*}{entailment} & \multirow{1}{*}{neutral} &  & P: yeah that's the World League &     \\
 &  & entailment & H: \hl{\textit{Though,}} that's the World League that you can join. & \cmark     \\
                               &                                & contradiction    & H: \textit{Though,} that's the World League that you can \hl{join}. & \cmark    \\ \midrule
\multirow{1}{*}{entailment} & \multirow{1}{*}{entailment} &  & P: Their ideas and initiatives can be implemented at the local and national levels. &     \\
 &  & contradiction & H: \hl{\textit{Indeed,}} locally and nationally, their ideas can be applied. & \cmark     \\
                               &                                & neutral    & H: \hl{\textit{Indeed,}} locally and nationally, their ideas can be applied. & \cmark    \\\bottomrule
                               
\end{tabular} 
}
\caption{Examples of the data staining experiment design. Each hypothesis is augmented with a prefix (\textit{italics)} conditioned on the stained class; \textit{entailment} with `Indeed,' and the others with `Though,'. For each foil, we test whether the stain is highlighted. If the stained class is the fact or the foil, we hypothesize that the stain \textit{should} be highlighted, thus it ``passes'' the test.
}
\label{tab:esnli-staining-ex}
\end{table*}
\begin{table}[t]
\centering
\resizebox{0.98\linewidth}{!}{%
  \begin{tabular}{llccc}
        \toprule \bf \multirow{2}{*}{Stained Class} & \multicolumn{3}{c}{\bf Stain Prefix} \\ 
        \cmidrule(lr){2-4}
        & \bf entailment & \bf contradiction & \bf neutral \\ \midrule
        entailment & Indeed, & Though, & Though, \\
        contradiction & Indeed, & No, & Indeed, \\
        neutral & And, & And, & Though, \\
        \bottomrule
\end{tabular}}
\caption{Three data stains used for MultiNLI, used as prefixes for the hypotheses.
For e.g., the entailment class is stained by prefixing all entailment examples with ``Indeed, '' while all other examples are prefixed with ``Though, ''.
Thus the model is encouraged to use the stain \textit{contrastively} only for or against the stained class.
The choice of prefix is intended to be semantically insignificant to the original text, though the stained dataset should nevertheless be considered synthetic by its nature, and not natural language.
}
\label{tab:mnli-stains} 
\end{table}
\begin{table}[]
\centering

\resizebox{0.98\linewidth}{!}{%
\begin{tabular}{llrrr} \toprule 
\bf \multirow{2}{*}{Stained Class}  & \bf \multirow{2}{*}{Fact} &  \multicolumn{3}{c}{\bf Foil}    \\ 
\cmidrule(lr){3-5}
                               &                       & \bf entailment & \bf contradiction & \bf neutral       \\ \midrule
\multirow{2}{*}{entailment}    & contradiction         & \textbf{.0202}     &       ---        & .0037   \\
                               & neutral               & \textbf{.1012}      & .0017       &   ---      \\ \midrule
\multirow{2}{*}{contradiction} & entailment            &      ---      & \textbf{.0172}        & .0059   \\
                               & neutral               & .0029      & \textbf{ .0668}        & ---        \\ \midrule
\multirow{2}{*}{neutral}       & entailment            &       ---     & .0065       & \textbf{.1058}   \\
                               & contradiction         & .0074     &        ---       & \textbf{.0791}  \\ \bottomrule
\end{tabular}}
\caption{Data staining results for foil ranking. The stained class emerges as the contrastive class for both possible facts, indicating that the model is indeed using the stain primarily in contrast to the stained class. 
}
\label{tab:mnli-data-staining-results}
\end{table}
\section{Sanity Checks via Data Staining in NLI} 
\label{app:staining}

We present a strss-test evaluation  to verify the validity of our contrastive framework via \textit{data staining} \cite{datastaining}.
It involves ``staining'' a training instance with a feature  (e.g. an inserted token) guaranteed to be useful for the task,  and then attempting to recover this feature via the analysis.\footnote{\citet{datastaining} introduced the stain by altering the label distribution. 
Our formulation is slightly different in that the stain manipulates the input.}
We modify data staining to evaluate contrastive explanations via introducing stains which are only \textit{contrastively} useful to the model; the ``stain'' is some feature useful to differentiate between a specific fact and a foil. 
The model is thus encouraged to exploit this feature in its decision making. 

Our stain is added to NLI hypothesis during training as shown in Table~\ref{tab:mnli-stains}.
The stain can be used by a model to perfectly distinguish a class from the others, i.e. the stain is \textit{contrastively useful} for or against only the stained class. 
See Appendix~\ref{app:staining} for illustrative examples with the stains. 


We analyze a \verb+RoBERTa-Large+ model fine-tuned on the stained MultiNLI train set. The entire MultiNLI dataset (train, dev-matched, dev-mismatched) was stained during the experiment, and we masked the stain for 10\% of our training data to ensure such examples are in distribution for the model, enabling us to use masked-stain examples in the analysis step.
We repeat our experiment thrice, considering one of the three NLI classes as a stain each time. 
In all three cases, the stained models achieve high predictive mismatched dev-set performance on the stained MultiNLI (above 97\% accuracy). \
This high performance is expected, and indicates that the models indeed exploit the stain features. 
\ye{why not 100\%?} \alon{why would it be 100?} \swabha{perhaps this can be attributed to the stain already occuring in the input in some way, and therefore not being too useful?}

To recover the stains on the MultiNLI dev-set using our methodology, we apply highlight masking interventions.
We define our candidate factor space to be all tokens in the hypothesis, and we expect the first token (the stain) to be the most contrastively important evidence when either the fact or the foil is the stained class. 
We report the accuracy of recovering the stain as the salient factor (ranking factors, c.f. \S\ref{subsec:measure}) when the fact or foil \swabha{foil as well? isn't that a little loose?} \alon{"for the fact/foil" and "against the fact/foil" are interchangeable logically, so I don't think this is possible to do otherwise} is the stained class, or recovering a non-stain word when the stained class is neither the fact nor the foil. \swabha{too loose a definition for True positives, but okay I guess.}
We perform the experiment for a random sample of 1000 test-set MultiNLI instances. 
The results are 98.45\% for the \textit{entailment}-stained model, 96.9\% for \textit{contradiction}, and 96.1\% for \textit{neutral}.\footnote{Since the model is \textit{not} guaranteed to make optimal use of the stain in every case, a high but sub-optimal accuracy performance is within expectations.} 
Table~\ref{tab:esnli-staining-ex} shows a demonstration of the data staining experiment.

Additionally, if the stain were to be treated as a highlight which we intervened on, followed by a contrastive intervention, and behavioral change, we would expect the stained class to be the most affected (ranking foils, c.f. \S\ref{subsec:measure}).
This is indeed the case, as shown in results in Table~\ref{tab:mnli-data-staining-results}.


\section{Hyp-Negation Concept in NLI} 
\label{app:negation-concept-ex}

In the Hyp-Negation experiments, as mentioned, we opted to produce counterfactuals of the hypothesis negation concept, instead of doing so via amnesic probing. 
Table~\ref{tab:negation-ex} contains examples of the 180 instances that we will make available online. 
\swabha{is this done in the repo?} \alon{yes}

\begin{table}[t]
\centering
\small
\resizebox{0.98\linewidth}{!}{%
  \begin{tabular}{lp{6.5cm}} 
  \toprule
Label   &  Text \\ \midrule
\multirow{2}{*}{Entailment} & P: A couple being romantic under the sunset. \\
 & H: A man and a woman touching each other  looking at the sunset. \\
 & H (\textit{negated)}: A man and a woman not far away from each other  looking at the sunset. \\ \midrule
\multirow{2}{*}{Entailment} & P: A man wearing yellow sneakers and a black vest is on a water board and hangs onto a waterski line that appears to be towed by a boat. \\
 & H: A man is moving on the water. \\
 & H (\textit{negated)}: A man is not sitting still on the water. \\ \midrule
\multirow{2}{*}{Neutral} & P:  man jumps in front of a palace in China. \\
 & H: The street is crowded. \\
 & H (\textit{negated)}: The street is not empty. \\ \midrule
\multirow{2}{*}{Neutral} & P: A football coach guiding one of his players on what he should do. \\
 & H: The coach knows how to play football. \\
 & H (\textit{negated)}: The coach does not need to learn how to play football. \\
  \bottomrule 
\end{tabular} 
}
\caption{Examples of counterfactuals used in the \textit{hyp-negation} concept experiment. The authors of this work produced the negated hypotheses to be used in the analysis. The original instances were taken from the SNLI dataset.
}
\label{tab:negation-ex}
\end{table}

\section{Highlight Ranking in BIOS} 
\label{app:highlight-ranking-bios}

Here, to adjust for space, we show some qualitative examples for the BIOS highlight ranking described in Section \ref{subsec:highlight-ranking-bios} presented in Table~\ref{tab:bios-ranking-factors-ex2}.

\begin{table}[t]
\centering
\ra{1.1}
\resizebox{0.98\columnwidth}{!}{%
  \begin{tabular}{llp{8cm}}
        \toprule Fact & Foil & Input with Highlights \\ 
        (prediction) & (\textbf{gold}) &  \\ \midrule
        \multirow{7}{*}{\rotatebox[]{90}{attorney}} & \multirow{2}{2cm}{no explicit foil} & Harris said the abuse had been inflicted by both “hands and items,” and, according \hl{to evidence,} since near the time of Kairissa’s arrival in Mt. Juliet.  \\
         & \multirow{2}{*}{\textbf{physician}} & \hl{Harris said} the abuse had been inflicted by both “hands and items,” and, according to evidence, since near the time of Kairissa’s arrival in Mt. Juliet.  \\ \midrule
        \multirow{20}{*}{\rotatebox[]{90}{{attorney}}} & \multirow{5}{2cm}{no explicit foil} & He has been involved in land transport for the past 13 years and has worked on various transport projects in Malta. He was appointed as chief officer for land transport within the Authority for Transport in Malta in 2010 where he was responsible for the \hl{regulation of} driver training, testing and licensing, vehicle registration, goods transport, and passenger transport. In 2015 he moved back to the private sector and took on the role of General Manager of the local bus company, Malta Public Transport with his main responsibility being to oversee the transformation of the public transport service.  \\
         & \multirow{5}{*}{\textbf{accountant}} & He has been involved in land transport for the past 13 years and has worked on various \hl{transport projects} in Malta. He was appointed as chief officer for land transport within the Authority for Transport in Malta in 2010 where he was responsible for the regulation of driver training, testing and licensing, vehicle registration, goods transport, and passenger transport. In 2015 he moved back to the private sector and took on the role of General Manager of the local bus company, Malta Public Transport with his main responsibility being to oversee the transformation of the public transport service.  \\ \midrule
         \multirow{20}{*}{\rotatebox[]{90}{{physician}}} & \multirow{5}{2cm}{no explicit foil} & She is a \hl{top medicine} student whose academic achievements are sweet fruit of her labor. All seemed well until she reached her junior internship year, when one of her patients died under her watch. She was publicly humiliated in the aftermath. Her closest friends and family tried to lift her spirits up, but to no avail. She thought she was a failure. All she felt was the immense pressure boiling inside of her, and she can no longer contain it. Thus, on one fateful night, on the rooftop of her apartment, she decides to end her misery by taking her own life.  \\
         & \multirow{5}{*}{\textbf{surgeon}} & She is a top medicine student whose academic achievements are sweet fruit of her labor. All seemed well until she reached her junior internship year, when one of her \hl{patients died} under her watch. She was publicly humiliated in the aftermath. Her closest friends and family tried to lift her spirits up, but to no avail. She thought she was a failure. All she felt was the immense pressure boiling inside of her, and she can no longer contain it. Thus, on one fateful night, on the rooftop of her apartment, she decides to end her misery by taking her own life. \\ \bottomrule
\end{tabular}}
\caption{
Qualitative examples for the BIOS highlight ranking described in Section \ref{subsec:highlight-ranking-bios}.
The top-1 results of ranking \textit{highlights} contrastively given a particular foil, compared to doing so generally (i.e., where the foil is all other classes together) for examples where the model made a mistake. 
We consider the space of highlights to be all unigrams and bigrams, and rank the space by the change in behavior (via difference of normalized logits) following a masking intervention on the highlight. 
In the example in the second row (for which the model mistakenly predicted \textit{physician}), the model is generally most affected by the bigram ``top medicine''. 
However, this is not a particularly useful feature to favor \textit{physician} rather than \textit{surgeon}, since surgeons also entail medicine studies; when we repeat the experiment in contrast to \textit{surgeon}, the top highlight changes to ``patients died'', indicating that this bigram is a better differentiator for those classes in the trained BIOS model.
}
\label{tab:bios-ranking-factors-ex2}
\end{table}

\end{document}